\documentclass{ecai} 

\usepackage{latexsym}
\usepackage{amssymb}
\usepackage{amsmath}
\usepackage{amsthm}
\usepackage{mathtools}
\usepackage{booktabs}
\usepackage{enumitem}
\usepackage{graphicx}
\usepackage{xcolor}
\usepackage{multicol,multirow}
\newcommand{\BibTeX}{B\kern-.05em{\sc i\kern-.025em b}\kern-.08em\TeX}

\begin{document}

%%%%%%%%%%%%%%%%%%%%%%%%%%%%%%%%%%%%%%%%%%%%%%%%%%%%%%%%%%%%%%%%%%%%%%%%

\begin{frontmatter}

%%% Use this command to specify your submission number.
%%% In doubleblind mode, it will be printed on the first page.

\paperid{5108} 

%%% Use this command to specify the title of your paper.

\title{Fin-Ally: Pioneering the Development of an Advanced, Commonsense-Embedded Conversational AI for Money Matters}
%%% Use this combinations of commands to specify all authors of your 
%%% paper. Use \fnms{} and \snm{} to indicate everyone's first names 
%%% and surname. This will help the publisher with indexing the 
%%% proceedings. Please use a reasonable approximation in case your 
%%% name does not neatly split into "first names" and "surname".
%%% Specifying your ORCID digital identifier is optional. 
%%% Use the \thanks{} command to indicate one or more corresponding 
%%% authors and their email address(es). If so desired, you can specify
%%% author contributions using the \footnote{} command.

\author[A]{\fnms{Sarmistha}~\snm{Das}\orcid{0000-0001-6608-0400}\thanks{Corresponding Author. Email: sarmistha\_2221cs21@iitp.ac.in}}
\author[A]{\fnms{Priya}~\snm{Mathur}}
\author[A]{\fnms{Ishani}~\snm{Sharma}} 
\author[A]{\fnms{Sriparna}~\snm{Saha}}
\author[B]{\fnms{Kitsuchart}~\snm{Pasupa}}
\author[C]{\fnms{Alka}~\snm{Maurya}}
\address[A]{Department of Computer Science and Engineering, Indian Institute of Technology Patna, India}
\address[B]{School of Information Technology, King Mongkut's Institute of Technology Ladkrabang, Thailand}
\address[C]{CRISIL Limited, India}

%%% Use this environment to include an abstract of your paper.

\begin{abstract}
% Conventional systems oversimplify financial concepts, such as limiting credit card use to shopping while neglecting their broader applications in bill payments and financial management, which increases the demand for commonsense-aware, human-like financial chatbots.
The exponential technological breakthrough of the FinTech industry has significantly enhanced user engagement through sophisticated advisory chatbots. However, large-scale fine-tuning of LLMs can occasionally yield unprofessional or flippant remarks, such as ``With that money, you're going to change the world,'' which, though factually correct, can be contextually inappropriate and erode user trust. The scarcity of domain-specific datasets has led previous studies to focus on isolated components, such as reasoning-aware frameworks or the enhancement of human-like response generation. To address this research gap, we present Fin-Solution 2.O, an advanced solution that 1) introduces the multi-turn financial conversational dataset, {\em Fin-Vault}, and 2) incorporates a unified model, \textit{Fin-Ally}, which integrates commonsense reasoning, politeness, and human-like conversational dynamics. \textit{Fin-Ally} is powered by COMET-BART-embedded commonsense context and optimized with a Direct Preference Optimization (DPO) mechanism to generate human-aligned responses. The novel \textit{Fin-Vault} dataset, consisting of 1,417 annotated multi-turn dialogues, enables \textit{Fin-Ally} to extend beyond basic account management to provide personalized budgeting, real-time expense tracking, and automated financial planning. Our comprehensive results demonstrate that incorporating commonsense context enables language models to generate more refined, textually precise, and professionally grounded financial guidance, positioning this approach as a next-generation AI solution for the FinTech sector.

\end{abstract}

\end{frontmatter}

%%%%%%%%%%%%%%%%%%%%%%%%%%%%%%%%%%%%%%%%%%%%%%%%%%%%%%%%%%%%%%%%%%%%%%%%

\section{Introduction}
The rapid proliferation of financial chatbots has reshaped customer service and operational efficiency within the financial sector. Market projections indicate a compound annual growth rate of 23.3\%, with the global chatbot market expected to reach \$15.5 billion by 2028~\cite{chatbot_statistics2024}. This growth is fueled by the increasing need for continuous customer support and cost-effective service delivery. With the expeditious growth of financial LLMs such as FinGPT \cite{liu2023fingpt}, FinLLAMA \cite{konstantinidis2024finllama}, InvestLLM \cite{yang2023investlm}, and FinROBOT \cite{zhou2024finrobot}, the financial sector has witnessed substantial enhancements in automation. Despite their expanding adoption, existing financial chatbots frequently fall short due to limited commonsense reasoning, which is crucial for accurately managing everyday financial interactions. Early chatbot implementations were often generic and lacked domain-specific contextual understanding, limiting their ability to resolve complex financial queries. For example, when asked about paying utility bills with a credit card, conventional chatbots might provide irrelevant responses such as, ``\textit{You can only use your card for shopping},'' underscoring a critical deficiency in contextual comprehension~\cite{fintech_chatbots2024, linkedin_chatbot_limits2024}. This persistent inadequacy has culminated in notable user dissatisfaction, underscoring the critical need for more sophisticated, context-aware financial assistants. While advanced decoder-only language models such as GPT-3.5~\cite{wang2023openchat} and its variants deliver impressive conversational fluency, they often falter in maintaining coherence across domain-specific, multi-turn dialogues. Moreover, large-scale models such as Bloomberggpt~\cite {wu2023bloomberggpt} and GPT-4o~\cite {hurst2024gpt}, despite their enhanced capabilities, remain proprietary, rendering their architectures inaccessible for fine-grained task-specific adaptation, which poses significant challenges for domain customisation. Compounding this, the sheer resource intensity of these models frequently results in a degradation of response quality over extended interactions; for example, a user query may elicit an unprofessional reply such as, "Okay! Let me spoon-feed you," which can erode user trust and diminish engagement. The demand for intuitive chatbots is underscored by the fact that 80\% of financial institutions regard AI-driven conversational agents as strategic assets for enhancing customer engagement~\cite{masterofcode_chatbot_stats2024}, while 43\% of U.S. banking clients prefer chatbots for resolving financial queries, reflecting growing reliance on automated financial services. 
% Nevertheless, delivering genuinely impactful user interactions necessitates transcending static, rule-based systems through the integration of commonsense reasoning and dynamic contextual understanding.

% This inadequacy has led to user dissatisfaction and exposed the necessity for more intelligent, context-aware financial assistants. Though advance decoder-only language models like GPT-3.5 \cite{} variatents  provide extraordinary experience during conversations, but lose context while domain-specific multi-turn conversations. Subsequently, large models like the BloombergGPT \cite{wu2023bloomberggpt} and GPT-4O \cite{} which are not opensource and getting hands on their architecture and make them task specific learnable is a big challenge. However, due to their large resource pertaining, often the LLMs tend to provide unprofessional responses after multiple turns, for instance, based on user's query the bot responds "okay! Let me spoon feed you", which often hurt the user's sentiment.

% The financial industry acknowledges the transformative potential of AI-driven chatbots, with 80\% of institutions recognizing them as strategic assets for enhancing customer engagement~\cite{masterofcode_chatbot_stats2024}. Additionally, over 43\% of U.S. banking customers prefer chatbots for resolving financial issues, reflecting a growing reliance on automated assistants~\cite{masterofcode_chatbot_stats2024}. However, achieving meaningful user interactions requires moving beyond static, rule-based responses by integrating commonsense reasoning capabilities.
\begin{figure}[hbt]    
\centering
    \includegraphics[width = 0.50\textwidth]{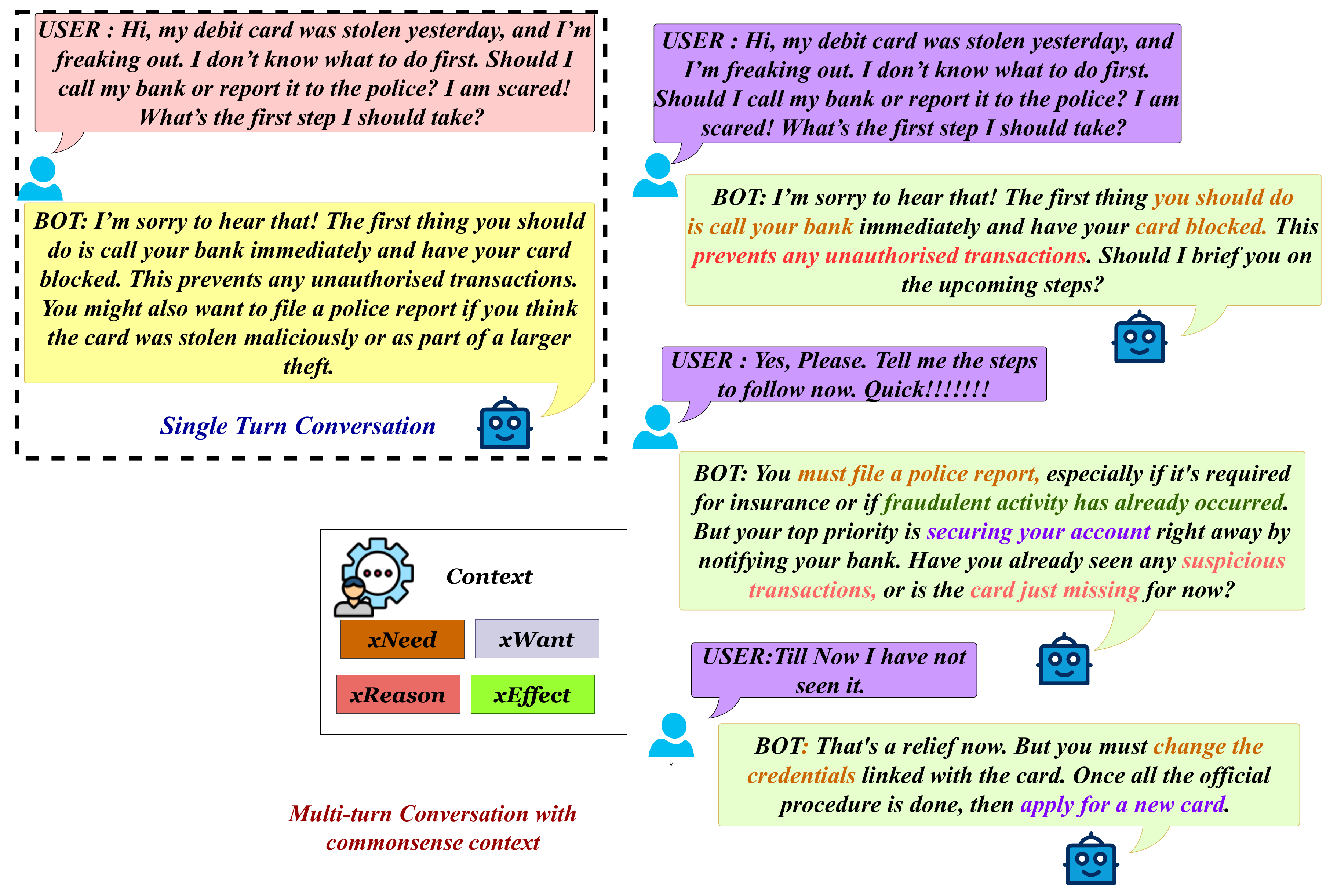}
    \caption{An advanced commonsense-aware financial chatbot capable of coherent multi-turn conversations. Where commonsense context enriches situational needs, reasons, wants, and effects}\label{intro_image}    
\end{figure}
\paragraph{Motivation:} Existing research addressing this task has predominantly concentrated on isolated components—such as dataset benchmarking~\cite{yuan2024finllms,xie2024finben}, reasoning enhancement~\cite{srivastava2024evaluating}, or improved alignment~\cite{ouyang2024ethical}—while overlooking critical dimensions such as  contextual comprehension and situational awareness, both of which are pivotal for navigating complex, real-world financial interactions. Conventional chatbots often generate contextually inappropriate responses; for example, when asked, ``\textbf{What should I do if my debit card is stolen?}'' a typical bot might respond with, ``\textbf{Visit the nearest ATM to check your balance}'' reflecting a severe lack of commonsense reasoning. To address these limitations, we present an ultimate solution, Fin-Solution2.O, that 1)
introduces the multi-turn financial conversational dataset \textit{\textit{Fin-Vault}},
and 2) incorporates a unified model, \textit{Fin-Ally}, which integrates commonsense reasoning, politeness, and human-like conversational dynamics.
Through \textit{Fin-Ally}, we aim to study the incorporation of each component's importance for dynamically interpreting user queries and providing contextually appropriate responses, such as recommending immediate card blocking and reporting in case of theft as described in Figure~\ref{intro_image}. The \textit{Fin-Vault} dataset is meticulously annotated with 1,417 multi-turn conversations, enriched with politeness categories to enhance \textit{ Fin-Ally}’s user-friendliness. To ensure responses are closely aligned with human preferences, the model leverages Direct Preference Optimization (DPO), enabling the generation of refined, contextually appropriate, and human-aligned conversational outputs.

% Recognizing the scarcity of specialized financial datasets, we developed \textit{FinVault}, a meticulously curated dataset of 850 annotated financial conversations. This dataset serves as a critical foundation for \textit{Fin-Ally}'s domain-specific training, enabling precise, adaptive learning and enhanced contextual accuracy. By bridging advanced AI technologies with domain expertise, \textit{Fin-Ally} represents a transformative step toward intelligent, responsive, and contextually aware financial assistance.

\paragraph{Research Objectives:}
\begin{enumerate}
\item To explore the integration of commonsense reasoning for generating coherent, context-aware multi-turn dialogues within the financial domain, effectively addressing routine monetary concerns.
% To investigate the efficacy of LLMs in facilitating contextually accurate and adaptive financial conversations. 
\item To rigorously assess the extent to which DPO enhances alignment with human conversational preferences, contributing to more natural and contextually appropriate responses.
\end{enumerate}

\paragraph{Contributions:}  
Our key contributions are as follows:
\begin{enumerate}
    \item Initially, we introduce \textit{Fin-Solution 2.O}, an integrated solution designed to address everyday financial tasks, comprising a unique, domain-specific dataset paired with a user-centric conversational AI assistant.
    
    \item Subsequently, we present \textit{Fin-Vault}, a pioneering financial dataset consisting of 1,417 meticulously annotated multi-turn dialogues spanning key financial sectors, including banking, credit and debit card management, insurance, and market investments.

\item Further, we introduce \textit{Fin-Ally}\footnote{\textbf{Fin-Ally: \url{https://github.com/sarmistha-D/Fin-Ally}}.}, the first commonsense-aware financial chatbot augmented with DPO, engineered to deliver contextually enriched and human-aligned conversations, establishing a new benchmark in AI-driven financial dialogue systems.

\item Moreover, we conduct comprehensive empirical evaluations under both zero-shot and fine-tuned settings, benchmarking multiple state-of-the-art LLMs to determine optimal architectures for task-specific financial chatbot deployment. Our findings provide an in-depth analysis of LLM performance, adaptability, and robustness within specialized financial dialogue contexts.

\end{enumerate}
% [5] https://www.ilo.org/media/313436/download

\section{Related Works}
The progression of financial chatbots mirrors the broader evolution of conversational AI, advancing from basic rule-based systems to sophisticated AI-powered agents. Early models lacked contextual understanding, but breakthroughs in machine learning and NLP have enabled personalised, context-aware interactions~\cite{semanticscholar_rulebased}. The latest architectures harness LLMS and Retrieval-Augmented Generation (RAG) frameworks, enabling real-time data access and hyper-personalised financial advisory services~\cite{ijert2024rag}. 
% Notable platforms, such as Sendbird and Kasisto’s KAI-GPT, exemplify this paradigm shift, providing secure, multilingual, and compliant chatbot solutions~\cite{sendbird2024ai}. 
Recent innovations spotlight domain-specific LLMs such as FinTral~\cite{bhatia2024fintral} and Open-FinLLMs~\cite{huang2024open}, which leverage multimodal inputs (text, images, tables) and multi-task fine-tuning to achieve state-of-the-art performance across diverse financial tasks. These models are meticulously engineered to enhance reasoning, decision-making, and cross-modal adaptability, often outperforming larger general-purpose models such as GPT-4-o \cite{hurst2024gpt} in specialised contexts. Subsequently, the open-source dataset diversity has emerged as a linchpin for model robustness. Resources such as InvestorBench~\cite{investorbench2024}, and ConvFinQA~\cite{chen2022convfinqa} offer rich, domain-specific corpora encompassing numerical reasoning, structured data interpretation, and compositional queries—crucial for training models to navigate real-world financial complexities.

Furthermore, the field is experiencing rapid advancements in reasoning-centric models, exemplified by Fin-R1~\cite{finr1_2024} and Fino1~\cite{qian2025fino1}, which leverage chain-of-thought (CoT) reasoning and a two-stage training paradigm—comprising Supervised Fine-Tuning and Reinforcement Learning—to achieve superior performance on financial reasoning benchmarks. However, a critical gap remains: the absence of a unified solution that integrates a multi-turn conversational dataset with embedded commonsense reasoning capabilities. From our research perspective, prior studies have predominantly focused on isolated components—whether dataset development, numerical reasoning, or personalized chatbot design—leaving a clear opportunity to develop a commonsense-aware conversational agent enhanced with preference optimization, capable of producing refined, human-aligned responses.

\section{Curation of Dataset}
% This section provides a concise overview of the dataset collection, preprocessing steps to ensure gold-standard quality, and the annotation process, highlighting key dataset characteristics (see Appendix~\ref{sec:dataset-appendix} for detailed dataset information).

\subsection{Data Collection}
To develop a robust financial advisory chatbot, we curated a domain-specific conversational dataset aimed at addressing the limitations of existing financial datasets, as detailed in Table~\ref{data-comp}. Most prior datasets are confined to structured financial data or short, isolated user queries, which restrict their usefulness in real-world advisory scenarios. To bridge this gap, we carried out an extensive analysis of online forums and community-driven platforms where individuals actively seek financial advice. Our focus encompassed key personal finance topics\footnote{\url{https://www.investopedia.com/terms/p/personalfinance.asp}} including credit vs. debit card use, loan management, taxation, EMIs, banking operations, stock investments, and bonds. As highlighted by Mitchell et al.~\cite{Lusardi2023importance}, mastery of these areas is critical for individuals to make informed financial choices, while active engagement in these domains enables financial institutions to drive product innovation and strengthen risk management strategies.

Data were collected from widely recognized platforms including Reddit’s r/personalfinance \footnote{\url{https://www.reddit.com/r/personalfinance/}} and r/financialplanning \footnote{\url{https://www.reddit.com/r/FinancialPlanning/}}, as well as specialized forums such as Bogleheads \footnote{\url{https://www.reddit.com/r/Bogleheads/}} and the Financial Wisdom Forum (FWF)\footnote{\url{https://www.financialwisdomforum.org/forum/}}. The data scraping process centered around high-impact keywords such as budgeting, credit cards, insurance, and investment strategies to ensure comprehensive coverage. Notably, we prioritized gathering financial queries that hold universal relevance across diverse demographic regions, particularly Europe and Asia, enhancing the dataset’s global applicability.

% Market Situation, Savings, Retirement Planning, Investment Strategies, Loans, Financial Goals, Peer Advice, Tax Returns, Emergency Funds. These platforms host discussions on topics including budgeting, retirement planning, insurance, and investment strategies, featuring responses from community members with varying expertise levels, thereby providing a rich mix of professional and peer-to-peer advisement.
% To develop a financial advisory chatbot that embodies commonsense embedded awareness, we curated a domain-specific conversational dataset addressing the limitations of existing financial datasets as showcased in Table~\ref{data-comp}, which often focuses on structured data or brief user queries. Our objective was to capture natural, multi-turn dialogues reflective of authentic user-advisor interactions. We conducted a comprehensive analysis of online forums and community-driven platforms where individuals seek financial guidance, such as Reddit's r/personalfinance and r/financialplanning, as well as specialized forums such as Bogleheads following key-words such as Budgeting, Credit Cards, Insurance,
% Market Situation, Savings, Retirement Planning, Investment Strategies, Loans, Financial Goals, Peer Advice, Tax Returns, Emergency Funds. These platforms host discussions on topics including budgeting, retirement planning, insurance, and investment strategies, featuring responses from community members with varying expertise levels, thereby providing a rich mix of professional and peer-to-peer advisement.

\begin{table}[ht]
\centering
\caption{Comparative Analysis of Contemporary Financial Datasets}

\label{data-comp}

\resizebox{\columnwidth}{!}{%
\begin{tabular}{cccc}
\toprule
\textbf{Dataset} & \multicolumn{1}{c}{\textbf{\begin{tabular}[c]{@{}c@{}}Diverse \\ Domains\end{tabular}}} & \textbf{Count} & \textbf{Multi-dialogue} \\ \midrule
TAT-QA~\cite{zhu2021tat}              & \checkmark                                                                                        & 16,552        & $\times$                                            \\ %\hline
FinQA~\cite{chen2021finqa}              & \checkmark                                                                                         & 8,281                               &  $\times$                                            \\ %\hline
FinanceBench~\cite{islam2023financebench}      & \checkmark                                                                                       & 10,231                              &  $\times$                                           \\ %\hline
FinTextQA~\cite{chen2024fintextqa}                             & \checkmark                                                                                         & 1,262                               &  $\times$                                         \\ 
\textit{Fin-Vault} (Ours)    & \checkmark                                                                                     & 1417                                 & \checkmark                                            \\ \bottomrule
\end{tabular}
}
\end{table}

\subsection{Data Sample Sanitization} 
To ensure data quality and contextual integrity, we adopted a two-phase validation strategy. During Evidence Confirmation, factual correctness was verified by cross-referencing financial claims against authenticated sources such as RBI (Reserve Bank of India) \footnote{\url{https://www.rbi.org.in/scripts/DataDefinition.aspx}} bulletins, SEBI (Securities and Exchange Board of India ) \footnote{\url{https://www.sebi.gov.in/}} advisories, and official banking portals. Entries with outdated or incomplete information were discarded, accounting for the frequent policy changes across Indian states. In the subsequent Relevance Evaluation, only data pertinent to current financial concerns—taxation changes, loan restructuring policies, and recent RBI mandates—was retained. To ensure comprehensive coverage of financial conversations, the dataset integrates a rich vocabulary across key categories: interrogative words (e.g., \textit{what}, \textit{how}, \textit{when}), financial domain terms (\textit{EMI}, \textit{interest}, \textit{KYC}, \textit{account}), conversational fillers (\textit{okay}, \textit{yeah}, \textit{please}), politeness phrases (\textit{could you}, \textit{please help}), action verbs (\textit{apply}, \textit{verify}, \textit{transfer}), emotion words (\textit{worried}, \textit{angry}, \textit{thanks}), discourse connectors (\textit{but}, \textit{then}, \textit{if}), and references to financial entities (\textit{PayPal}, \textit{Chase}, \textit{Google Pay}). These lexical classes underpin semantic understanding, sentiment analysis, and task-oriented dialogue modeling, enabling the \textit{Fin-Vault} corpus to support context-aware, goal-driven financial advisory systems.
% To ensure data quality, we focused on threads with substantial up-votes, positive community engagement, and active moderation. Our preliminary screening of 100 threads, in collaboration with financial domain experts, revealed that users often candidly express financial concerns, providing detailed context. These discussions frequently include follow-up questions and clarifications, offering valuable insights into conversational dynamics.
\begin{table}[]
\centering
% \scriptsize

\caption{Over-all statistical details of proposed \textit{Fin-Vault} Corpus}
\label{fin-stat}
\scalebox{0.79}{
\begin{tabular}{l|r|r|r}
\toprule
Statistic                                   & Count  & Domain & Count   \\
\midrule
Vocabulary size                             & 3398 & Stock & 453   \\ 
Avg. no. of tokens per User                 & 10.80 & Investment & 185 \\
Avg. no. of sentences per User              & 1.07  & Personal Finance &141  \\
Avg. no. of tokens per BOT's response       & 41.18 & Banking & 150 \\ 
Avg. no. of sentences per BOT's response    & 2.93 & Loan & 114  \\ 
Avg no. of words presents per Conversation  & 145.33 & General Finance & 52 \\ 
Avg no. of tokens presents per Conversation & 145.33 & Credit Card & 66\\ 
No. of unique bigrams in queries                     & 17477 & Tax & 102 \\ 
% No. of unique bigrams in bot response                     &   \\ 
Total No. of unique trigrams in  dataset              &  139204 & Trading & 69 \\ 
Avg no. queries in Conversations                  & 234.88 & others & 85 \\ \bottomrule
\end{tabular}}
\end{table}

\begin{figure}[ht]    
\centering
    \includegraphics[width = 0.8\columnwidth]{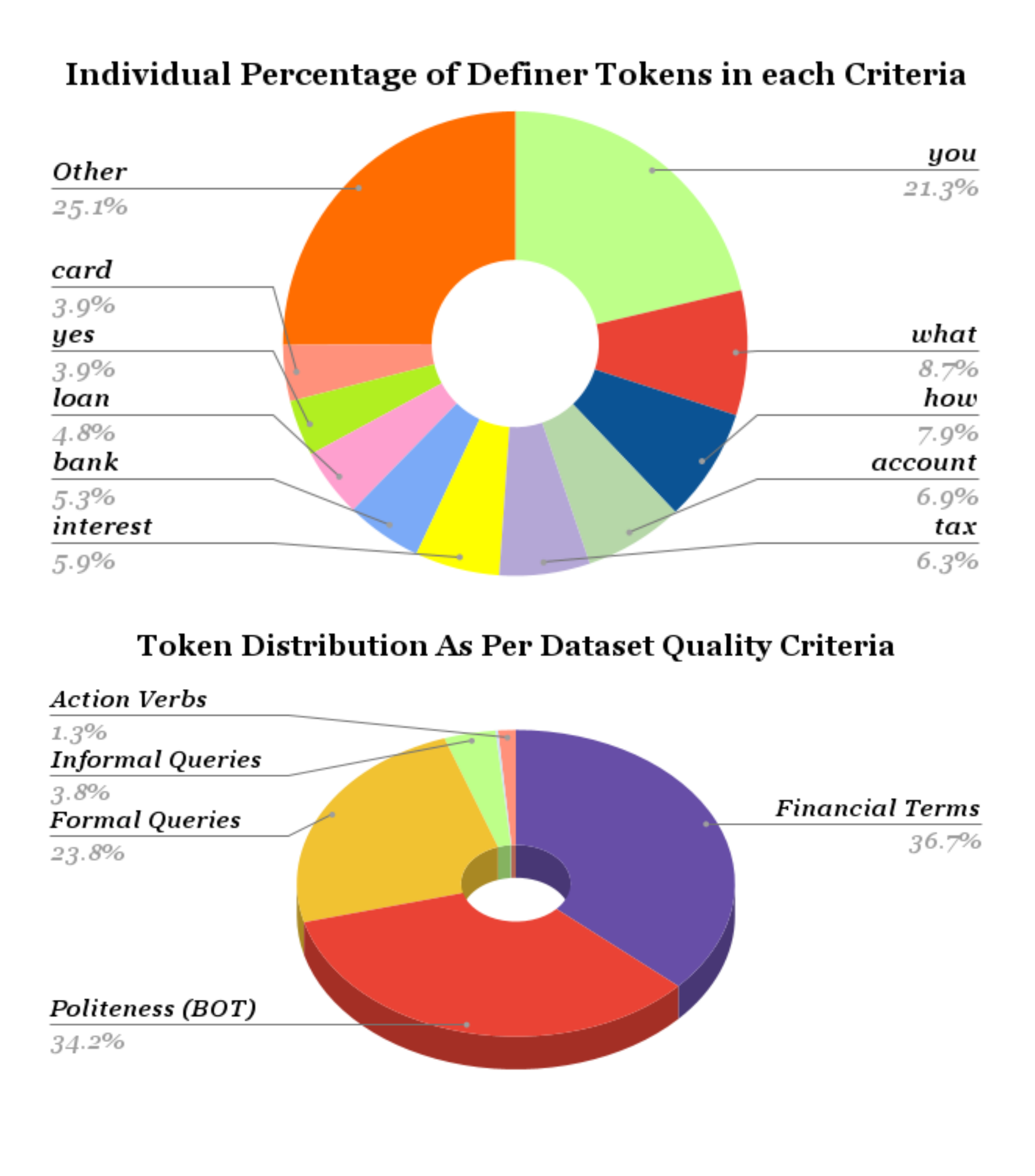}
    \caption{Fine-grained token-level statistical illustration in the curated \textit{Fin-Vault} dataset.}\label{Wordcloud}
\end{figure}

From publicly available sources, we retrieved 1,800 high-engagement financial advisory threads using official APIs and web scraping tools. Adhering to strict authentication, rate limits, and data integrity measures, we prioritised 10 key financial domains with the politeness category. Within these domains, we curated conversations featuring multiple Q\&A exchanges covering various scenarios (advice-seeking, clarification requests, context provision) from diverse demographic region as mentioned in Table \ref{fin-stat}. Each conversation underwent stringent quality checks, ensuring factual accuracy, grammatical coherence, and clarity. Ultimately, 1417 conversations—comprising over 4,006 utterances. The final dataset includes 2,743 queries classified as globally relevant, alongside 290 queries specific to the USA, 542 to India, 19 to the UK, 55 to Canada, 60 to Australia, and 297 to Europe. This balanced dataset captures both general and region-specific financial concerns, forming a solid foundation for culturally adaptive, context-aware advisory systems that authentically simulate real-world financial dialogues with textual references, numerical data, and essential stylistic nuances. Figure \ref{Wordcloud} describes the linguistic backbone of the proposed dataset.
\subsection{Annotation}
As underscored in Table \ref{data-comp}, multi-turn conversational datasets within the financial domain remain notably limited. To address this gap, we introduce \textit{Fin-Vault}, a rigorously curated, user-centric dataset comprising structured user-bot dialogues that exemplify conversational fluency and incorporate detailed politeness annotations. Advisory-oriented content is extracted from sanitised conversational threads, meticulously analysed, and reconstructed into coherent multi-turn exchanges (\(\geq\) 3 turns) to mirror authentic financial consultations. In light of the absence of explicit demographic metadata, annotators conduct thorough source verification and systematically annotate demographic attributes, categorising dialogues by regions such as Europe, Asia, and Africa. To position \textit{Fin-Vault} as a robust benchmark for financial conversational systems, each dialogue is additionally classified according to politeness levels: “Polite,” “Impolite,” or “Neutral.” This transformation from raw advisory content to refined user-bot dialogues is executed through a stringent two-phase annotation protocol, ensuring precision, coherence, and domain-specific integrity.\\
The annotation process was executed in two distinct phases. We assembled a six-member team comprising three undergraduate students in finance and economics as junior annotators, one doctoral candidate in computer science as the senior annotator, and two seasoned financial industry advisors from reputable institutions as expert validators. All team members brought strong domain expertise and advanced linguistic proficiency in finance-specific discourse.

\subsubsection{Phase-1}
In the initial phase, the senior annotator collaboratively produced 100 annotated samples to serve as benchmark references for the junior annotators. To initiate the process, the junior annotators were first provided with 50 raw conversational samples, along with comprehensive annotation guidelines. To ensure annotation quality, each sample was evaluated using E-FAIR metrics—Engagement, Fluency, Adequacy, Information Preservation, and Readability—rated on a 1–5 scale, with each criterion assigned a score of 1 if present.
% Each conversation was sourced from a distinct financial advisory forum post, featuring a dyadic exchange between a ``user'' seeking financial guidance and an ``assistant'' that represented our expert system, constructed from community-generated commentary. 
During this preliminary stage, any uncertainties related to the annotation procedure or instances of lexical ambiguity were promptly addressed. These iterative discussions and clarifications continued until the annotated samples consistently achieved E-FAIR ratings in the moderate-to-good range (3-4). Subsequently, the junior annotators were given an additional 50 seed conversational samples, and their work was evaluated against the established reference annotations produced by the primary annotators. The financial expert annotators then assessed these new annotations, assigning E-FAIR scores on a scale from 1 to 5. Once the annotated samples attained an average score exceeding 3, the second and final phase of the annotation process commenced.

\subsubsection{Phase-2} 
In the subsequent phase, the remaining 1,317 samples were partitioned into three subsets, each assigned to a distinct junior annotator. Throughout this stage, the annotators were explicitly instructed to operate in isolation, ensuring the independence and integrity of their annotation processes. Upon completing their respective subsets, the junior annotators engaged in a structured cross-validation procedure, critically evaluating and assigning quality E-FAIR ratings (1 - 5) to one another's annotated samples. The senior annotator then conducted a thorough review of these results, performing any necessary manual refinements to uphold the highest quality standards.

The dataset comprises user-agent exchanges, each consisting of one to three question-answer pairs, carefully curated to span a broad range of financial topics, as illustrated in  Figure~\ref{Wordcloud}.

\section{Methodology}
\subsection{Problem Definition} 
\begin{figure*}[ht]    
    \centering
    \includegraphics[width = 0.77\textwidth]{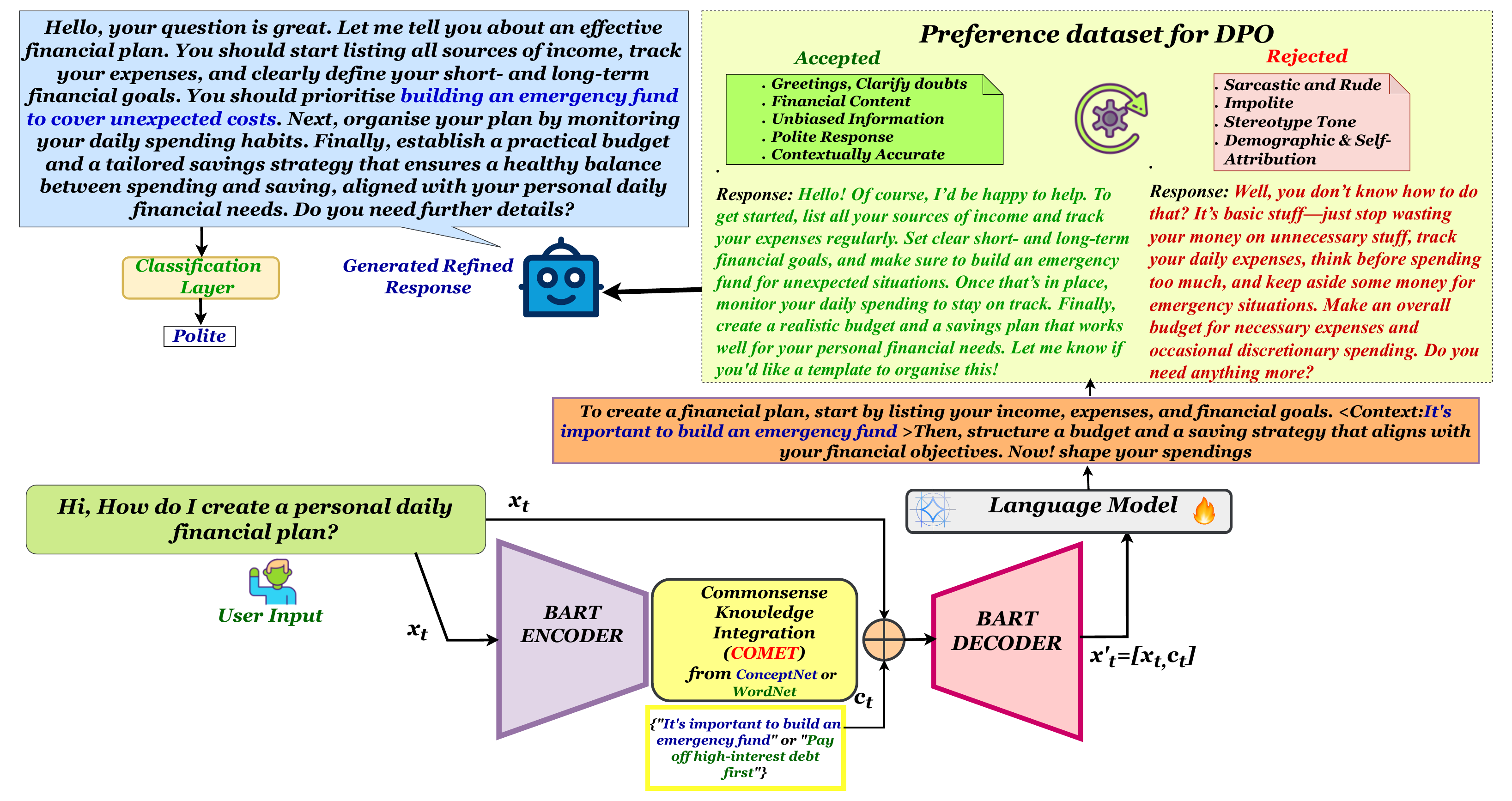}
    \caption{Proposed \textit{Fin-Ally} Architecture: A Commonsense-Aware Financial Chatbot with DPO Configuration}\label{architecture}    
\end{figure*}

The goal of this research is to develop a conversational generative model that, given user input texts $t$, generates interactive and contextually relevant responses $y$ by incorporating commonsense-aware context $c$. Formally, for a given conversation $i$ with conversation ID $id_i$, the input is a sequence of user queries $t = (t_1, t_2, \ldots, t_n)$, and the corresponding model's task is to predict the sequence of bot responses $y = (y_1, y_2, \ldots, y_n)$ conditioned on both the user inputs and the dynamic commonsense context $c$. The model learns from a dataset $D = \{(id_i, t, y, c)\}$ where each conversation $id_i$ contains multiple paired utterances (user inputs $t_i$ and bot responses $y_i$), to generate coherent, contextually aware, and relevant responses that improve over time as the system processes more conversations. Figure~\ref{architecture} illustrates the architecture of the proposed framework, comprising three sequential phases: Phase 1 entails the generation of commonsense-aware contextual embeddings; Phase 2 involves supervised fine-tuning of the language model, further enhanced through Direct Preference Optimization (DPO); and Phase 3 incorporates a politeness classification module to ensure the generated responses adhere to conversational norms.

\subsection{Generating Commonsense-Aware Financial Contexts Using COMET-BART}

To enhance financial commonsense reasoning, COMET-BART~\cite{hwang2021comet} processes a user’s financial query represented as a token sequence \(\mathbf{X} = [x_1, x_2, \dots, x_n]\), where each \(x_i\) denotes an input token. This sequence is tokenized using the BART tokenizer~\cite{lewis2019bart} into token IDs \(\mathbf{Z} = [z_1, z_2, \dots, z_m]\), which are passed through the BART encoder to obtain contextual embeddings \(\mathbf{H} = [h_1, h_2, \dots, h_m]\), with each \(h_i \in \mathbb{R}^d\) capturing semantic context. Simultaneously, external commonsense financial knowledge \(\mathbf{K}\) is retrieved from structured sources such as ConceptNet~\cite{speer2017conceptnet}, linking financial concepts through triples such as ("mutual funds", "UsedFor", "diversifying investments") and ("retirement", "RelatedTo", "long-term financial planning"). These triples are converted into natural language statements (e.g., "Mutual funds are used for diversifying investments"), embedded using Sentence-BERT \cite{reimers2019sentence}, and filtered via cosine similarity against the query embedding (threshold: 0.7).

The selected knowledge $\mathbf{K}$ is integrated with the encoder's contextual embeddings $\mathbf{H}$ via a fusion mechanism, yielding enriched representations $\mathbf{H}' = \text{Fuse}(\mathbf{H}, \mathbf{K})$. The decoder then generates output tokens $\mathbf{Y} = [y_1, y_2, \dots, y_p]$ autoregressively, computing the probability of each token conditioned on prior outputs $\mathbf{Y}_{<i}$ and $\mathbf{H}'$ as:
$$
P(y_i \mid \mathbf{Y}_{<i}, \mathbf{H}') = \text{Decoder}(y_{i-1}, \mathbf{H}').
$$
This process continues until the sequence $\mathbf{Y}$ is complete, which is then decoded into human-readable text $\mathbf{R} = \text{Decoder}(\mathbf{Y})$. For example, the generated response might be: "Investing in mutual funds is a common strategy for diversifying your portfolio, which supports long-term goals such as retirement planning." Here, $\mathbf{R}$ is contextually enriched by both the user query $\mathbf{X}$ and external knowledge $\mathbf{K}$, ensuring that the advice is financially informed and commonsense-aware.

 An example of COMET-BART contexts begins with a user query: ``I keep hearing about stocks and bonds; what's the difference, and which one should I consider first?'' The model infers several contextual elements, including the user's reasoning (xReason: I want to invest in stocks and buy bonds), desire (xWant: to buy a bond), prerequisite state (xNeed: to be interested in investing), intent (xIntent: to know the difference), and related concept (RelatedTo: I want to buy a stock). Using these inferred contexts, the system generates a response: ``Stocks represent ownership in a company, while bonds are loans to companies or governments that pay interest. Stocks offer higher risk and potential reward; bonds are steadier with lower returns. Your choice should reflect your risk tolerance.''

% first generates implicit knowledge $c_t$ from a given user query $x_t$, capturing financial risks, investment strategies, or economic principles. This retrieved commonsense knowledge is concatenated with the original query to form an enriched input $x'_t = [c_t; x_t]$, which is then tokenized using the BART tokenizer and encoded bidirectionally to produce contextual embeddings $H$. Instead of direct augmentation, the decoder attends to both $H$ and retrieved knowledge $c_t$ through multi-head self-attention, generating the output sequence $S$ autoregressively as $P(s_i | S_{<i}, H, c_t) = \textrm{Softmax}(W_q Q + W_k K + W_v V)$, where $Q, K, V$ are the query, key, and value matrices for context integration, respectively. The final output $R$ ensures a coherent, knowledge-grounded financial response, making chatbot recommendations contextually rich and financially informed.

\subsection{Supervised Fine-Tuning of LLM for Financial Advice}
We fine-tune a large language model (e.g., Gemma2-9B) using supervised learning on a dataset consisting of financial queries and expert-annotated responses. The dataset is structured as input-output pairs $(x'_t, y)$, where $x'_t$ is the commonsense-augmented user query, and $y$ is the corresponding expert-validated financial advice. This fine-tuning process enables the chatbot to generate responses aligned with professional financial guidance.

\subsection{Optimization Using DPO}
% During inference, we observed that the model exhibited various biases, including sarcastic or dismissive tones, condescending remarks, and assumptions based on gender or race. For instance, some responses implied that women are more risk-averse and should invest conservatively, while men should opt for riskier investments. Similarly, there were instances where users from Western countries received high-value investment recommendations, whereas users from developing regions suggested more conservative financial products. Additionally, self-attribution bias was detected, where the model confidently attributed financial success to its own advice while neglecting market unpredictability. 

To improve more human-aligned generated responses $y$ further, we employed DPO by constructing a preference dataset $\mathcal{D}_{\text{pref}}$ with three response types: Preferred Response ($y^+$), which is polite, expert-validated, and context-aware; Rejected Response ($y^-$), containing impolite such as stereotyping, sarcasm, or misleading claims; and \textit{Chosen Response}, the human-annotated gold standard. This structured approach enables DPO to refine outputs, ensuring neutrality and reliability in financial advisory responses.

These responses were generated using GPT-3.5 Turbo~\cite{wang2023openchat} and validated by two financial experts to ensure fairness, accuracy, and ethical considerations in financial recommendations. The model was optimised using the DPO loss function, which enhances its preference for unbiased responses by maximising the preference gap between $y^+$ and $y^-$:
\begin{multline}
\mathcal{L}_{\text{DPO}}(\theta) = 
- \sum_{\mathclap{\substack{(x_t, y^+, y^-) \\ \in \mathcal{D}_{\text{pref}}}}}
\log \sigma \left( \beta \cdot \left( \log P_{\theta}(y^+ \mid x_t)\right.\right. \\[-4ex]
\left.\left. - \log P_{\theta}(y^- \mid x_t) \right) \right),
\end{multline}
where $\beta$ is a hyperparameter that controls the preference gap, ensuring that the chatbot prioritizes responses aligned with human expert advice.

Finally, we evaluate the chatbot's financial advice generation by comparing its responses to gold-standard expert advice. This methodology ensures that the chatbot provides neutral, well-informed, and personalised financial advice while actively mitigating impoliteness that could otherwise affect user trust and financial decision-making.

\subsection{Politeness Classification}
To analyse generated response quality, we fine-tuned a RoBERTa-based classifier~\cite{liu2019roberta} for politeness detection. Given a tokenized utterance $\mathbf{U} = [u_1, \dots, u_m]$, RoBERTa encodes contextual embeddings $\mathbf{H}_p = [h_1, \dots, h_m]$, from which the \texttt{[CLS]} token vector $\mathbf{r} \in \mathbb{R}^d$ is extracted and projected as $\mathbf{o} = \mathbf{W}_p \mathbf{r} + \mathbf{b}_p$, where $\mathbf{W}_p \in \mathbb{R}^{C \times d}$ and $\mathbf{b}_p \in \mathbb{R}^C$. This yields logits over $C=3$ classes—\textit{Polite}, \textit{Neutral}, and \textit{Impolite}—with the final probabilities obtained via a softmax function, enabling robust politeness classification.

% To complement response generation, we fine-tuned a RoBERTa-based classifier~\cite{liu2019roberta} to predict politeness levels. Given a tokenized utterance $\mathbf{U} = [u_1, u_2, \dots, u_m]$, RoBERTa encodes it into contextual embeddings $\mathbf{H}_p = [h_1, h_2, \dots, h_m]$, where $h_j \in \mathbb{R}^d$. A pooled vector $\mathbf{r} \in \mathbb{R}^d$ (e.g., the [CLS] token) summarises the utterance and is projected through a linear layer:
% \[
% \mathbf{o} = \mathbf{W}_p \mathbf{r} + \mathbf{b}_p
% \]
% where $\mathbf{W}_p \in \mathbb{R}^{C \times d}$ and $\mathbf{b}_p \in \mathbb{R}^c$ yield logits over $C$ politeness classes ($C \in \{\text{Polite}, \text{Neutral}, \text{Impolite}\}$). Subsequently, a softmax function produces class probabilities, enabling nuanced tone classification by leveraging RoBERTa’s contextual encoding for robust politeness detection.

\section{Experimental Results and Discussion}
This section thoroughly details the experimental framework, offering a comprehensive exposition of baseline models, an in-depth comparative performance evaluation, and a critical analysis of the proposed \textit{Fin-Ally} model alongside the \textit{Fin-Vault} dataset. Our research addresses the following research questions (RQ):
\begin{itemize}
    \item \textbf{RQ1:} How does incorporating commonsense contextual reasoning influence the model's effectiveness, particularly in multi-turn financial conversations?
    \item \textbf{RQ2:} Does DPO truly optimize generated responses and enhance performance measurements in financial advisory contexts?
    \item \textbf{RQ3:} What are the broader societal implications of the proposed system and dataset, and how well can its solutions scale across a variety of NLP tasks, both within the financial sector and beyond?
\end{itemize}

In this study, we considered several state-of-the-art LLMs for their suitability in task-specific financial chatbot applications. The models considered include FLAN-T5-Base~\cite{https://doi.org/10.48550/arxiv.2210.11416}, Vicuna-7B~\cite{zheng2023judging}, GPT-3.5 Turbo~\cite{wang2023openchat} FinanceConnect-13B~\cite{ceadar_2023}, Mistral-7B~\cite{jiang2023mistral}, Llama-3-8B~\cite{llama3modelcard} and proposed Gemma-2-9B~\cite{gemma_2024} embedded \textit{Fin-ALLy}. 

\paragraph{Experimental Setup:}
To ensure consistent and unbiased experimental evaluation, we employed uniform configurations across the seven language models, training them on the curated financial dataset using finetuning and with and without context-aware settings. Experiments were conducted utilizing 4-bit precision for optimized memory efficiency. The models were trained with identical hyperparameters: temperature = 1, top\_k = 5, do\_sample = True, max\_target\_length = 1024, and a learning rate of $1\times10^{-4}$. We utilized the Adam optimizer with a weight decay of 0.01, ensuring stable convergence and consistent performance metrics.

\paragraph{Evaluation Metrics:} Model performance was assessed using four core evaluation metrics: BLEU~\cite{lin-och-2004-orange}, ROUGE~\cite{lin-2004-rouge}, BERTScore~\cite{zhang2019bertscore}, and METEOR~\cite{banerjee2005meteor}. BLEU and ROUGE measure $n$-gram overlap, emphasizing lexical correspondence between generated and reference texts, while BERTScore leverages contextual embeddings from pre-trained language models to evaluate semantic congruence. METEOR further balances precision and recall while incorporating synonym matching, enhancing flexibility in handling linguistic variation. These metrics provide a comprehensive evaluation framework, capturing both surface-level accuracy and deep semantic coherence.

\begin{table*}[t]
\caption{The ablation studies between popular fine-tuned language model with commonsense embedded context consideration. (R1--RL): ROUGE variants; (B1--B3): BLEU variants; BSP: BERTScore Precison;  BSR: BERTScore Recall;  BSF: BERTScore F1; P: Politeness and MS: Meteor Score. WC stands for without context setting. Bold values indicate the highest scores, while underlined values denote the second highest, for context and DPO+Context setting.}
\label{ablation}
\resizebox{\textwidth}{!}{
\begin{tabular}{cccccccccccccc}
\toprule
\multirow{2}{*}{Model} & \multirow{2}{*}{Settings} & \multicolumn{12}{c}{Metrics} \\ \cmidrule{3-14}

 & & {R1} & {R2} & {RL} & {B1} & {B2} & {B3} & {B4} & {BSP} & {BSR} & {BSF1} & {P} & {MS} \\ \midrule

Vicuna-7B& \multirow{7}{*}{WC}& 23.30& 7.71 & 19.39& 34.71& 10.72& 5.32 & 3.04& 88.14& 86.20& 87.14& 31.37& 15.86\\

FinanceConnect-13B && 25.72& 8.38 & 21.98& 38.67& 12.23& 7.32 & 4.54& 88.59& 86.82& 88.23& 32.43& 17.31\\

FLAN-T5-Base && 23.22& 7.65 & 19.55& 35.03& 11.03& 5.21 & 2.71& 88.26& 86.24& 87.23& 30.13& 16.07\\

GPT-3.5 Turbo&& 31.92& 12.57& 24.2 & 22.63& 14.51& 8.82 & 8.81& 81.96& 82.39& 82.52& 30.68 & 20.84\\

LLaMA3-8b&& 24.87& 9.44 & 21.44& 21.43& 13.28& 7.16 & 4.19& 88.68& 86.55& 87.59& 34.51& 18.49\\

Mistral-7b && 23.31& 8.54 & 20.05& 38.69& 12.58& 6.48 & 3.58& 88.87& 86.31& 87.55& 33.67& 16.43\\

Gemma2-9b&& 23.79& 8.83 & 20.49& 37.94& 12.95& 6.8& 3.84& 88.58& 86.42& 87.47& 33.23& 16.83\\
\midrule
Vicuna-7B& \multirow{7}{*}{Context} & 23.76& 8.14 & 20.01& 36.19& 11.56& 5.41 & 2.84& 88.48& 86.23& 87.33& 34.27& 16.38\\

FinanceConnect-13B && 25.29& 9.31 & 21.27& 39.93& 13.12& 7.01 & 3.91& 88.73& 86.01& 87.15& 35.1 & 18.66\\

FLAN-T5-Base && 23.51& 7.88 & 20.08& 35.87& 11.13& 5.12 & 2.74& 88.22& 86.06& 87.12& 32.6 & 16.14\\

GPT3.5 Turbo && \textbf{\underline{34.30}} & \textbf{\underline{19.18}} & \textbf{\underline{30.71}} & \textbf{29.60} & \textbf{\underline{22.34}} & \textbf{15.65} & \textbf{\underline{17.07}} & \textbf{\underline{89.56}} & \textbf{\underline{87.63}} & \textbf{\underline{88.24}} & \textbf{\underline{33.23}} & \textbf{28.09} \\

LLaMA3-8b&& 24.31& 8.82 & 20.68& 37.66& 12.23& 6.12 & 3.27& 88.45& 86.43& 87.41& 36.78& 16.97\\

Mistral-7b && 22.03& 8.68 & 19.41& 39.89& 13.73& 6.69 & 3.61& 89.12& 86.04& 87.53& 36.67& 15.41\\

Gemma2-9b&& 26.12& 9.31 & 19.23& 28.31& 14.76& 8.21 & 4.11& 88.23& 86.18& 81.27& 38.23& 19.83\\
\midrule
LLaMA3-8b& \multirow{3}{*}{DPO+WC}& 30.03& 8.74& 21.33& 17.51& 8.72 & 4.12 & 2.19& 87.14& 86.43& 87.36& 44.26& 19.64\\

Mistral-7b && 34.37& 12.69& 25.89& 19.47& 11.08 & 7.15 & 4.83& 84.54& 83.17 & 86.09& 41.48& 20.10\\

Gemma2-9b&& 34.87& 15.56& 28.16& 22.02& 13.35& 8.99 & 6.5 & 88.64& 88.79& 88.69& 42.11& 24.75\\
\midrule
LLaMA3-8b& \multirow{3}{*}{DPO+Context} & 32.06& 10.86& 23.16& 18.76& 9.81 & 6.03 & 4.18& 88.04& 88.34& 88.17& 45.26& 21.64\\

Mistral-7b && 36.39& 14.94& 27.92& 21.49& 13.1 & 9.05 & 6.81& 89.52& 88.7 & 89.09& 43.29& 24.22\\

Gemma2-9b&& \textbf{36.77} & \textbf{21.56} & \textbf{32.16} & \textbf{\underline{29.02}} & \textbf{23.35} & \textbf{\underline{15.19}} & \textbf{18.05} & \textbf{89.94} & \textbf{89.79} & \textbf{88.69} & \textbf{52.02} & \textbf{\underline{27.75}} \\ \bottomrule
\end{tabular}
}
\end{table*}
%%%%%%%%%%%%%%%%%%%%%%%%%%%%%%%%%%%%%%%
\begin{table*}[t]
\centering
\caption{Error analysis of selected baseline models, with red highlights indicating extraneous contextual elements introduced in the responses.}

\label{Error}
% \resizebox{\textwidth}{!}{%
\resizebox{\textwidth}{!}{
\begin{tabular}{p{3cm}l}
\hline
User Question                   & How does a stock operate when it is listed between two exchanges?  \\ \hline
Ground Truth                    & \begin{tabular}[c]{@{}l@{}}While a stock listed on multiple exchanges isn't directly influenced by one another, price differences can create arbitrage opportunities that traders\\   exploit. Being listed on two exchanges increases a stock's liquidity.\end{tabular}            \\ \hline
Gemma2-9b + WC + DPO & \begin{tabular}[c]{@{}l@{}}When a stock is listed between two exchanges, it means the stock is traded on both exchanges simultaneously. This can happen when {\color{red}{a company}}  \\ {\color{red}{expands its operations or seeks to reach a wider investor base}}. For instance, a company might list on both the New York Stock Exchange (NYSE)\\   and the London Stock Exchange (LSE). This dual listing allows investors in both regions to trade the stock, increasing liquidity and accessibility.\\The stock price may vary slightly between the exchanges due to differences in trading hours, market conditions, or investor demand. However,\\  {\color{red}{the stock is still considered the same security, and the prices typically converge}} over time.\end{tabular} \\ \hline
Llama3-8b +  WC + DPO  & \begin{tabular}[c]{@{}l@{}}If a stock trades between multiple exchanges, it ensures wider availability to investors across {\color{red}{different geographic locations}}.\end{tabular}                                                       \\ \hline
Mistral-7b + WC + DPO & \begin{tabular}[c]{@{}l@{}}When a stock is listed on two exchanges, it operates similarly on both. The exchange where the {\color{red}{stock is traded first will execute the transaction,}}\\ {\color{red}{and the other exchange will act as a backup.}} The transaction price is determined by the exchange where the trade occurs.\end{tabular}\\ \hline
\end{tabular}}
\end{table*}

\subsection{Derived Discussion}
This section comprehensively analyses the experimental results, highlighting their implications while systematically addressing the aforementioned research questions through critical evaluation and in-depth interpretation.

\paragraph{Answer to RQ1 (Commonsense as a context impact):} The ablation study presented in Table~\ref{ablation} illustrates that incorporating commonsense contextual reasoning noticeably enhances model performance across multi-turn financial conversations. Models with contextual information consistently outperform their WC (Without Context) counterparts, as seen in GPT-3.5 Turbo, which improves across all metrics. {\em Fin-ALLy} (Gemma2-9B with DPO+Context) excels in generating refined financial advisory responses, achieving top scores in all metrics with evident performance gap against DPO+with out context setting. The integration of commonsense-aware contextual reasoning enhances fluency, relevance, and user engagement. For instance, when asked, ``Should I invest in stocks or bonds?'' a COMET-BART-enhanced model not only explains their differences but also leverages inferred commonsense knowledge (e.g., xIntent: ``to know the difference'', xWant: ``to buy a bond'') to provide personalized investment suggestions based on risk preference, financial goals, and prior user context. 

\paragraph{Answer to RQ2 (DPO with Context Impact):} 
It is evident from Table~\ref{ablation} that DPO effectively enhances response generation in financial advisory contexts by refining model alignment and optimizing performance metrics. DPO-trained models consistently outperform their non-DPO counterparts, achieving higher ROUGE, BLEU, and BERTScore metrics, indicating improved coherence and informativeness. {\em Fin-Ally} (Gemma2-9B with DPO+Context) achieves top scores, outperforming non-DPO baselines. DPO-trained models also show higher Politeness and Meteor Scores, enhancing fluency and user engagement. DPO fine-tuning aligns model preferences, ensuring contextually coherent and user-friendly financial advice. Incorporating DPO eliminates biases, focusing on individual risk tolerance and investment goals rather than demographic stereotypes. For instance, when asked, ``Should I invest in stocks or bonds?'' a DPO-trained model provides neutral, well-structured advice based on financial principles, avoiding self-attributed statements such as ``Young men prefer stocks while women prefer bonds.''

\paragraph{Answer to RQ3 (Dataset and Proposed System's Impact)}
The proposed \textit{Fin-Vault} dataset plays a crucial role in enhancing AI-driven financial assistance by introducing commonsense-aware, multi-dialogue, and diverse-domain financial conversations. Unlike existing datasets, \textit{Fin-Vault} captures real-world financial contexts, enabling models to generate more contextually relevant and user-centric responses. Its structured yet adaptable nature allows fine-tuning across various NLP applications beyond finance, including legal advisory, customer support, and healthcare. By leveraging DPO and contextual augmentation, the dataset improves model robustness in multi-turn, reasoning-intensive interactions. DPO integration enhances the scalability for both small Gemma-2-9B and  large GPT-3.5 Turbo language models. However, ensuring bias mitigation and computational efficiency remains a key challenge. \textit{Fin-Vault} lays the foundation for more inclusive, explainable, and adaptable commonsense-aware NLP models across industries.

\begin{figure}[hbt]    
\centering
    \includegraphics[width=0.8\columnwidth]{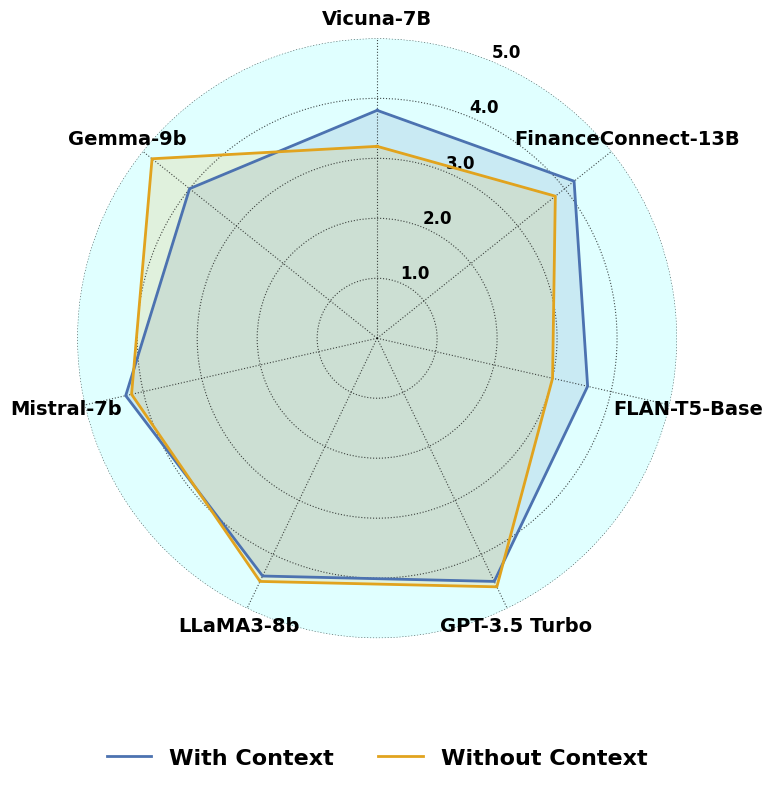}
    \caption{Comparison of human evaluation scores among the prominent models with or without commonsense context injection.}\label{Human-eval}    
\end{figure}

\subsection{Analytical Discussion}

\paragraph{Qualitative Analysis:} Evidently, adding commonsense context significantly improves model accuracy across all cases, making responses more aligned with financial regulations and best practices. DPO further refines this by enhancing clarity, user-friendliness, and structured reasoning. Models without commonsense context tend to provide vague or cautious responses, often stating that fees depend on regulations or policies, while models with context generate more confident, precise, and structured explanations. Among the models, Vicuna-7B and FLAN-T5-Base benefit from context but remain somewhat generic, whereas FinanceConnect-13B and GPT-3.5 Turbo perform better by incorporating disclosure requirements and policy nuances. LLaMA-3 8B and Mistral-7B see substantial improvements with context and DPO, providing clearer, more actionable insights. However, the proposed Gemma2-9b is the best performer due to its superior readability, balanced detail, and structured delivery. 

\paragraph{Human Evaluation:} Figure~\ref{Human-eval} illustrates the results of the human evaluation conducted on 140 samples, where expert annotators assessed model responses based on five criteria: fluency, adequacy, consistency, retention of financial terms, and readability. Each criterion was rated on a 1-5 scale. 
The evaluation highlighted that most models perform better when provided with additional context, highlighting their reliance on contextual information for improved outputs. However, an exception is Gemma2-9b, which demonstrates strong performance even without context while maintaining high-quality responses when context is injected. This suggests that the proposed Gemma2-9b configured \textit{Fin-Ally} is inherently robust in understanding and generating responses, relying less on explicit context compared to other models. In contrast, some models exhibit a significant drop in performance without context, emphasizing their dependence on additional information. Gemma2-9b's ability to sustain high-quality responses across both settings showcases its adaptability and strong generalization capabilities, making it a reliable choice for context-aware reasoning tasks.
\paragraph{Error Analysis:} The error analysis in Table~\ref{Error} highlights that Gemma+ without context + DPO provides the most accurate response, capturing liquidity, market-driven price variations, and price convergence but lacks mention of arbitrage opportunities. LLaMA3-8b+ without context + DPO is overly generic, failing to address key financial dynamics. Mistral+ without context + DPO introduces a critical error by misrepresenting dual-listed stocks as sequentially traded rather than independently influenced by arbitrage. While Gemma2-9b configured \textit{Fin-Ally} excels in clarity and coverage, incorporating arbitrage mechanics would enhance its accuracy, whereas LLaMA3-8b lacks depth and Mistral-7b misinterprets exchange operations.
\section{Conclusion \& Future Work}
This research introduces an ultimate financial solution, called Fin-Solution2.O, followed by \textit{Fin-Vault} dataset and {\em Fin-Ally}, a commonsense-aware financial chatbot fine-tuned on Gemma2-9b with DPO, designed for everyday financial conversations. Experiments using fine-tuning models with and without integrated commonsense context and DPO revealed that incorporating commonsense reasoning significantly enhances chatbot performance, especially when explicit commonsense contextual information is available. Notably, both the smaller parameter Gemma-2-9B and the larger GPT-3.5 Turbo models delivered superior results, as confirmed by automated metrics and human evaluations. These findings demonstrate that integrating commonsense reasoning with DPO enhances contextual understanding and advances intelligent financial conversational agents.
\par Our future endeavours include extending this work by developing a multimodal financial conversational agent capable of supporting multi-turn interactions with image inputs. We also aim to investigate politeness as a feedback signal to further improve response quality and user alignment.
% \section{Limitation}
% \begin{enumerate}
%     \item Due to authentication and privacy concerns, most money-related conversations are not publicly accessible, limiting the availability of comprehensive financial dialogue datasets.
%     \item Generic models such as Vicuna-7B and FLAN-T5-Base are not trained on finance-specific datasets, which explains the performance limitations observed even after fine-tuning financial tasks.
%     \item Despite FinanceConnect-13B being pretrained on a large-scale dataset and the COMET model leveraging the ATOMIC~\cite{rajpurkar2018know} dataset, the model occasionally generates repetitive or contextually overlapping outputs for relations such as xNeed and xWant, reflecting context hallucination. For instance, consider the following user query: ``\textit{Hey, My spouse and I are considering buying our first home. How much of a down payment should we aim for?}'' In this case, both xNeed and xWant are predicted as ``\textit{to save money},'' leading to redundancy and a lack of nuanced differentiation between the two relations.
%     \item Securing sufficient GPU resources for conducting all experiments remains a significant challenge, impacting the scalability and efficiency of model training and evaluation.
% \end{enumerate}

\section{Acknowledgement}
This work is a joint collaboration between the Indian Institute of Technology Patna and CRISIL Data Science Limited. We extend our sincere gratitude to Tannu and Jeel Doshi for their invaluable contributions as dataset annotators.

\bibliography{mybibfile}

\end{document}